# STATISTICAL TESTING ON GENERATIVE AI ANOMALY DETECTION TOOLS IN ALZHEIMER'S DISEASE DIAGNOSIS

Rosemary He

(Affiliation) Department of Computer Science, Graduate School of University of California, Los Angeles
Rosemary068@g.ucla.edu

Supervisor: Ichiro Takeuchi

(Affiliation) Graduate School of Engineering, Nagoya University
takeuchi.ichiro.n6@f.mail.nagoya-u.ac.jp

## ABSTRACT

Alzheimer's Disease is challenging to diagnose due to our limited understanding of its mechanism and large heterogeneity among patients. Neurodegeneration is studied widely as a biomarker for clinical diagnosis, which can be measured from time series MRI progression. On the other hand, generative AI has shown promise in anomaly detection in medical imaging and used for tasks including tumor detection. However, testing the reliability of such data-driven methods is non-trivial due to the issue of "double dipping" in hypothesis testing. In this work, we propose to solve this issue with selective inference and develop a reliable generative AI method for Alzheimer's prediction. We show that compared to traditional statistical methods with highly inflated p-values, selective inference successfully controls the false discovery rate under the desired α level while retaining statistical power. In practice, our pipeline could assist clinicians in Alzheimer's diagnosis and early intervention.

## 1. INTRODUCTION

Alzheimer's disease (AD), the most prevalent cause of dementia, is difficult to diagnose due to our limited understanding of disease mechanism and interpersonal variability between symptoms and progression[1]. While AD can only be formally diagnosed through autopsy, alternative biomarkers are used to assess clinical diagnosis. There are three classes of potential biomarkers for AD: β-amyloid, tau and neurodegeneration or neuronal injury[1]. Compared to the first two classes, neurodegeneration can be measured noninvasively and observed in time series magnetic resonance (MR) images. On the other hand, deep learning approaches for medical diagnosis and classification have shown potential to outperform traditional statistical models[2]. In the field of medical imaging, generative AI (genAI) has been applied extensively for tasks including anomaly detection and disease classification[3]. However, the reliability of such data-driven methods remains questionable due to its black-box nature and the issue of "double dipping", where the hypothesis is selected and validated on the same dataset[4]. While traditional statistical methods follow the assumption that result statistics are inherently independent of the selection criteria under the null hypothesis[4], data-driven methods, including deep learning, break this assumption by using the same dataset for training and validation. As a result, statistical inference becomes invalid with a high false discovery rate (type I error).

### 1.1 Related Works

In medical imaging, pathology such as anomalous growth (tumors) or traumatic injury can be viewed as deviations from the norm. Therefore, quantifying such deviation can be used to separate true anomalous signal from noise in imaging, improving classification accuracy. The main approach for anomaly detection with brain MRIs are semi-supervised, where the model is trained on the set of healthy patients only[16]. During inference, the model estimates the "healthy" alternative of the diseased image, and anomaly analysis is performed on the reconstruction error between the ground truth and prediction. While most published applications of such methods on brain imaging focus explicitly on tumors (which is one image), we apply the same setting to study anomalous neurodegeneration rate in Alzheimer's patients (which is progressional and estimated using two images).

Hypothesis testing in data-driven methods including machine learning and deep learning often face the issue of "double dipping"[4], where the same data is used to formulate and validate the hypothesis. As traditional statistical testing follows the assumption that a hypothesis is predetermined and independent from the dataset, "double dipping" will lead to selection bias, resulting in a high false discovery rate (FDR) and an invalid test. Selective inference (SI) is a statistical framework that corrects such bias by considering a conditional test. Previously, the SI framework has been applied to assess deep learning methods for medical segmentation[23], saliency map[24], anomaly detection in VAE[6] and diffusion[25] models.

### 1.2 Contribution

In this work, we propose to apply selective inference, a statistical testing framework for data-driven hypothesis[5], [6], to assess the reliability of a genAI based anomaly detection tools for AD neurodegeneration progression. We first develop a conditional variational autoencoder (CVAE) model for anomaly detection in Alzheimer's cohort, then apply selective inference to test the significance of its predictions. While both tasks have been studied in their respective fields, we list two novel contributions of our work: 1. the first to apply selective inference to assess the validity of genAI based diagnosis tool in Alzheimer's, and 2. novel approach to study neurodegeneration rate via estimating optical flow from paired MRIs.

## 2. METHODS

### 2.1 Data Preparation

We obtain our dataset from the Alzheimer's Disease Neuroimaging Initiative (ADNI) database[7] and the Open Access Series of Imaging Studies (OASIS) database[8], [9]. ADNI (adni.loni.usc.edu) is led by Principal Investigator Michael W. Weiner, MD. The primary goal of ADNI is to measure the progression of mild cognitive impairment (MCI) and early Alzheimer's disease. First, we include subjects from ADNI 1, 2 and 3 studies, as well as OASIS-3 who have at least 2 MPRAGE scans available at least one year apart. For patients with more than 2 scans, we take the first and last scans as the pair. In recent deep learning applications, disease progression can be modeled as changing pixel values in time series medical images[10], [11], [12], [13]. Under this model, we propose to estimate the atrophy rate via optical estimation, i.e. by tracking pixel movement in the pair of images over an arbitrary amount of time[14], [15]. For each pair of images, we estimate the velocity of change for each pixel, where positive values (light pixels) indicate growth and negative values (dark pixels) indicate atrophy. Sample visualization of optical flow estimation is shown in Fig. 1, where growth in the ventricles (due to atrophy in surrounding white matter) is observed. We note here that though both the MRI and optical flow are 3D (80x80x80 pixels), we take the middle coronal slice (80x80 pixels) as 2D input into the model due to computational limitations. Lastly, we standardize the images across the cohort.

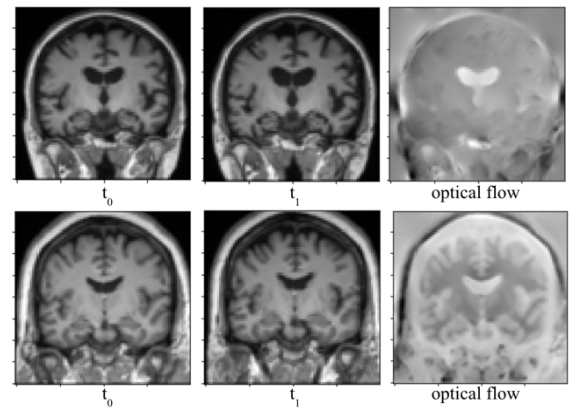

Fig. 1 sample visualization of optical flow estimation.

In addition to the images, we add two conditional variables to the model: age at the first scan and time difference between the pair of MRIs. We standardize both to have mean 0 and variance 1. For disease labels, we chose binary outputs for simplicity and group patients into healthy or diseased. We note here that ADNI groups patients into 6 categories: cognitive normal, subjective memory complaint, early mild cognitive impairment, mild cognitive impairment, late mild cognitive impairment and AD. Since OASIS and ADNI have different conventions for disease status, we keep consistency by grouping OASIS subjects into cognitive normal, cognitive impairment and AD, and assign them a value of 0, 3 and 5. To increase data quality, we take only the cognitive normal cohort as healthy, and late mild cognitive impairment (label of 5) and AD (label of 6) as diseased. After initial filtering and processing, the cohort consists of 888 healthy and 110 diseased individuals.

### 2.2 Anomaly Detection

#### 2.2.1 Conditional variational autoencoder

In this section, we give a brief overview for those unfamiliar with autoencoder models and their extensions. Autoencoders are a class of deep learning methods that learn a low-dimensional representation of high-dimensional data such as medical imaging[17]. Autoencoders often follow a U-net[18] architecture, where an encoder projects high dimensional data down to a latent space with lower dimensions, and a decoder learns to map the latent space representation back to its original input. Though there is a latent space, autoencoders are deterministic as its latent distribution is unknown, making inference difficult and prompting the need for variational autoencoders (VAE)[19].

VAE models assume the following sample generation: a sample $x$ is generated from a distribution conditioned on the latent space z, where z follows $p_\theta(z)$ and x follows $p_\theta(x|z)$. However, estimating the posterior distribution of x is intractable in practice. To address this issue, a more tractable distribution, $q_\Phi(z|x)$, is used to approximate the true posterior. VAEs can be trained to sample from unknown distributions using the variational lower bound of the log-likelihood, which can be written as follows[19]:

$$L(\theta, \Phi, x) = - D_{KL}(q_\Phi(z|x) \| p_\theta(z)) + E_{q_\Phi(z|x)}[\log p_\theta(x|z)] \quad (1)$$

, where $D_{KL}$ is the Kullback–Leibler divergence[20] that measures how similar two distributions are, and $E_{q_\Phi(z|x)}$ measures the sum of squared reconstruction error where we assume $p_\theta(x|z)$ is Gaussian with mean predicted by the model and variance of 1. We use the common choice that $p_\theta(z)$ is multivariate standard normal, with no learnable parameters. Conditional VAE (CVAE) models are an extension to the VAE and include an additional variable $y$ as the conditional variable[21]. While the mathematical framework remains the same, CVAEs are used widely in medical applications as conditional variables offer additional information and can improve parameter estimation.

### 2.2.2 Model architecture

To reduce computational time of selective inference testing, we construct a relatively simple CVAE as in Fig. 2. The encoder consists of 3 blocks, where each block contains one convolutional and one max pooling layer; the decoder consists of 3 blocks, one convolutional and one up-sampling layer with skip connections, as illustrated in grey in Fig. 2. As the selective inference procedure has high computational requirements, we keep the model relatively simple with a latent space size of 10 and 128 channels at the deepest.

### 2.2.3 Model training

Following a common training and inference scheme of anomaly detection, we train on healthy subjects only and keep all 110 AD subjects for selective inference later. Within 888 healthy subjects, we take 600 as the training set, 100 as the test set, 100 for selective inference and 88 for variance estimation used in selective inference. We split the subjects at random and train the model with the Adam optimizer[22], learning rate of 1e-5 and for 1000 epochs with early stopping.

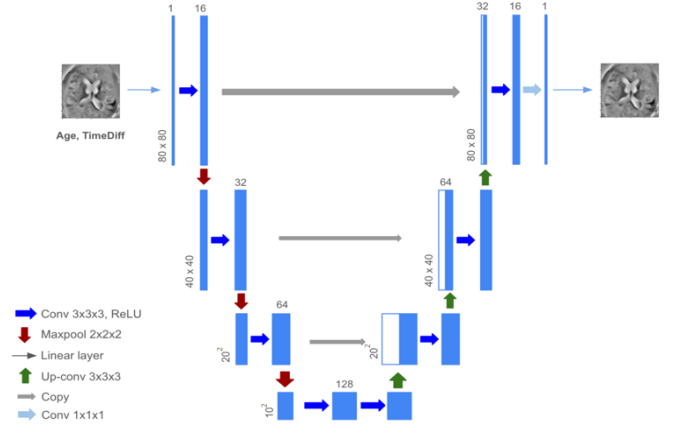

Fig. 2 Model architecture.

### 2.2.4 Anomaly detection pipeline

After the model is trained with only healthy subjects, we describe the steps for anomaly detection as follows:

1. Given the image of an Alzheimer's subject, the CVAE predicts the "healthy" alternative image
2. Take the difference between ground truth and prediction to find reconstruction error per pixel
3. Binarize the error by applying a threshold where a pixel $X_i = 1$ if abs($X_i$)>threshold and 0 otherwise
4. Obtain the anomaly mask

For the threshold, we take the 95[th] percentile of the reconstruction error for the healthy test set. To improve mask quality, we focus on the ventricle regions only.

## 2.3 Selective Inference

### 2.3.1 Statistical test

Here we follow the same set-up to assess our CVAE model as [6]. Let an optical flow image X be described as a n-dimensional vector,

$$X = (X_1, X_2, \ldots X_n) = \mathbf{s} + \boldsymbol{\varepsilon}, \boldsymbol{\varepsilon} \sim N(\mathbf{0}, \Sigma) \quad (2)$$

, such that $\mathbf{s} \in \mathbb{R}^n$ represents the vector of true signals and $\boldsymbol{\varepsilon} \in \mathbb{R}^n$ is the noise following normal distribution. Let the CVAE anomaly detection pipeline be represented as a black-box function A such that

$$A: \mathbb{R}^n \ni X \mapsto A_x \in 2^{[n]} \quad (3)$$

, where $2^{[n]}$ is the power set of [n] and $A_x$ is the anomaly mask obtained by steps in 2.2.4.

To quantify the reliability of detected anomaly regions, we test the difference between the true signal of pixels in the abnormal region $\{s_i\}_{i \in A_x}$ and normal region $\{s_i\}_{i \in A^c_x}$, where

$A^c_x$ is the complement to the anomaly region. We define the null and alternative hypothesis as:

$$H_0: \frac{1}{|A_X|} \sum_{i \in A_X} s_i = \frac{1}{|A^c_X|} \sum_{i \in A^c_X} s_i$$
$$H_1: \frac{1}{|A_X|} \sum_{i \in A_X} s_i \neq \frac{1}{|A^c_X|} \sum_{i \in A^c_X} s_i \quad (4)$$

, and the test statistics T(X) as

$$T(X) = \frac{1}{|A_X|} \sum_{i \in A_X} X_i - \frac{1}{|A^c_X|} \sum_{i \in A^c_X} X_i = \eta^\top X \quad (5)$$

$$\eta = \frac{1}{|A_X|} \mathbf{1}_{A_X} - \frac{1}{|A^c_X|} \mathbf{1}_{A^c_X} \quad (6)$$

In the naïve setting, the p-value can be calculated as

$$P_{naive} = P_{null}(|T(X)| \geq T(x)|) \quad (7)$$

, where X is a random vector and x is the observed image. Under this assumption, $P_{naive}$ can be calculated from the normal distribution $T(X) \sim N(0, \eta^T \Sigma \eta)$. However, it does not consider the fact that the anomaly region obtained is dependent on each input, resulting in a high false positive rate and rendering the test invalid[6]. On the other hand, SI calculates the p-value of T(X) conditional on the observation, i.e. $T(X)|\{A_X=A_x\}$. By conditioning on the hypothesis selection event, SI considers the rarity of observation from the subset of hypothesis in which an abnormal region is detected instead of the entire space. For details on calculating the statistics, we follow the same procedure as in [6] and use the SI package from [26].

## 3. RESULTS

First, we show the validity of SI under this setting by constructing a dataset with 1000 random noise images. Under the null hypothesis, the p-values of a valid test follows a uniform distribution, as seen in Fig. 3. The naïve p-values are heavily skewed to the left while the p-values from SI (selective p-value) are uniformly distributed as expected.

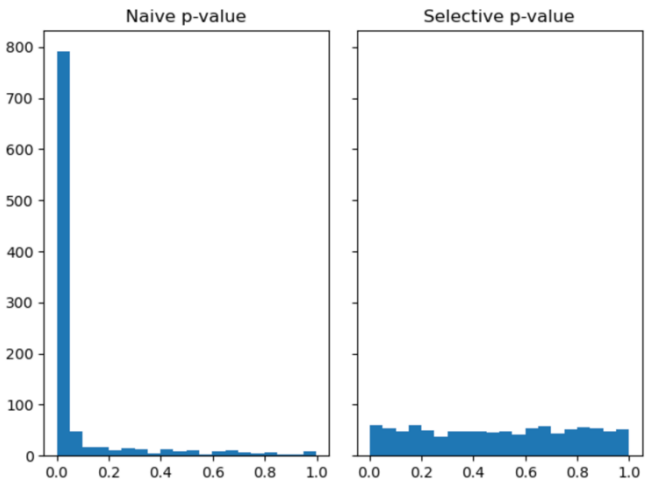

Fig. 3 p-value distribution under the null.

In addition to the naïve p-value, we calculate the adjusted p-values under the Bonferroni correction[27] as an alternative method for multiple testing. In Table 1, we compare the false discovery rate between the naïve, Bonferroni and SI, where the naïve and Bonferroni p-values are calculated using α=0.05 and SI values are calculated using α=[0.01, 0.05, 0.1]. As expected, the naïve method has a very high false discovery rate while Bonferroni has the lowest. We note here that SI controls the FDR well around the desired α at three different thresholds. In Table 2, we compare the power between the methods. We note power for the naïve method is not meaningful due to the high FDR. SI outperforms Bonferroni by a large margin and as α increases, the power of SI increases as expected from the FDR-power trade-off. We note here that SI's power is still not ideal, we suspect the input data may not provide a strong enough signal and will be a next step for this work.

Table 1. FDR comparison between methods

| Held-out normal | Reject the null | Failed to reject | FDR |
|---|---|---|---|
| Naïve | 74 | 26 | 0.74 |
| Bonferroni | 0 | 100 | 0 |
| SI [α=0.05] | 6 | 94 | 0.06 |
| SI [α=0.01] | 2 | 98 | 0.02 |
| SI [α=0.1] | 10 | 90 | 0.1 |

Table 2. Power comparison between methods

| Alzheimer's cohort | Reject the null | Failed to reject | FDR |
|---|---|---|---|
| Naïve | 93 | 17 | 0.85 |
| Bonferroni | 7 | 103 | 0.06 |
| SI [α=0.05] | 36 | 74 | 0.33 |
| SI [α=0.01] | 12 | 98 | 0.11 |
| SI [α=0.1] | 51 | 59 | 0.46 |

Lastly, we include some sample cases for visualization and comparison between the naïve and SI p-values in Fig. 4. For the extreme case where there is little to no anomaly detected, both the naïve and SI failed to reject the null. On the other hand, when there are large amounts of signals, both methods correctly identify the anomaly case of Alzheimer's. For cases in the middle, however, the naïve method overestimates the anomaly and the SI method underestimates the anomaly, which can be reflected by the high FDR of naïve and low power of SI.

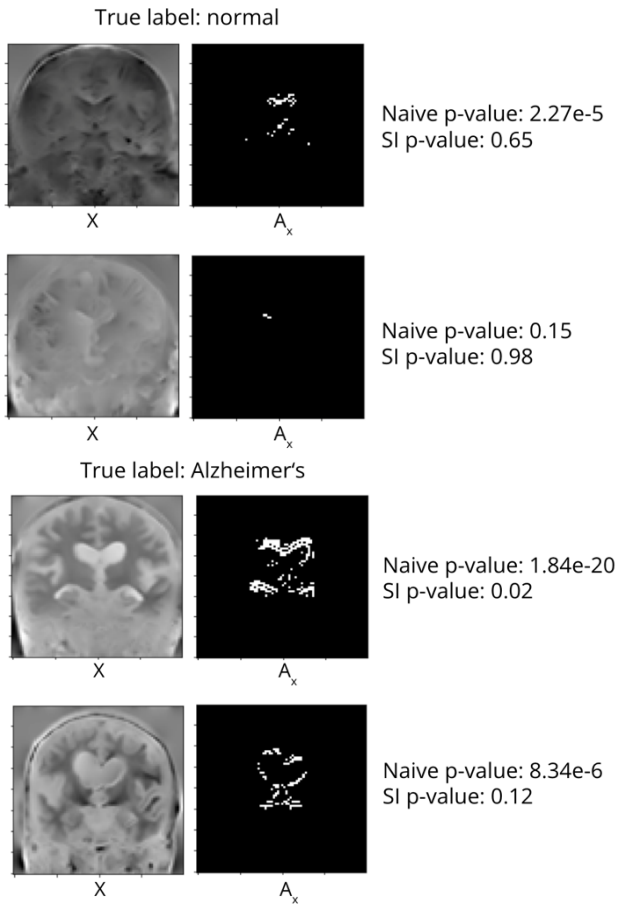

Fig. 4 Sample visualization and p-values.

## 4. DISCUSSION

In this work, we applied the selective inference framework onto anomaly detection for neurodegeneration in Alzheimer's patients to develop a reliable method for disease prediction. We show that selective inference controls the FDR well under three different α thresholds when the traditional hypothesis test fails. In addition, we show that compared to the more conservative Bonferroni method, SI has more power in detecting true anomalies, though SI's power should to be further improved upon for practical utilization. We suspect a few things could be done to improve the power. First, the dataset can be further improved and cleaned. Optical flow[14] estimates pixel change well over the short period of time, but does not work well for longer time elapses. For some patients, the time difference between MRIs could be on the magnitude of years and optical flow does not track this change well and does not provide enough signals. As alternative options for estimating the atrophy rate between pairs of images are explored, a better dataset could lead to better results. Secondly, hyperparameter choices can impact SI results. Finding a more appropriate anomaly threshold, instead of using the 95th percentile of the normal reconstruction error from the healthy test set, may help improve SI's power. Lastly, we took a 2D slice from a 3D image due to computational limitations, but in doing so greatly reduces the amount of information provided. An extension of the pipeline from 2D to 3D will incorporate more morphological changes in the brain and can potentially improve results. Another limitation of this work is that only two standard methods were compared, alternative solutions to double dipping such as data thinning[28] were not included. While all limitations listed above are also future directions of this work, this work provides an additional tool for anomaly detection using generative AI that can be statistically quantified reliability with just one dataset. As genAI tools face challenges including interpretability and reliability issues with integration into clinical practice, we hope this work serves as a starting point where more robust methods could be built upon for trustworthy AI-assisted clinical care.

## ACKNOWLEDGEMENTS

This work is supported by the Japan Student Services Organization (JASSO) scholarship and Nagoya University JUACEP program.